\pdfoutput=1

\documentclass[11pt,hyphens]{article}

\usepackage{acl}
 \usepackage{booktabs}
\usepackage{subfiles}
\usepackage{tabularx}
\usepackage{times}
\usepackage{latexsym}
\usepackage{multirow}
\usepackage{adjustbox}
\usepackage{tabularray}
\usepackage{float}
\usepackage{graphicx}
\usepackage{epstopdf}
\graphicspath{figures/}
\usepackage{titlesec}
\usepackage[T1]{fontenc}
\newcommand{\revision}[1]{\textcolor{blue}{#1}}

\definecolor{myGreen}{RGB}{34, 139, 34}

\definecolor{myTeal}{RGB}{0, 128, 128}
\newcommand{\final}[1]{\textcolor{myTeal}{#1}}
\newcommand{\finalZ}[1]{\textcolor{myTeal}{#1}}

\usepackage[utf8]{inputenc}
\usepackage{hyperref,nameref}
\usepackage{soul}
\usepackage{graphicx}
\newif\ifusecolor
\usecolorfalse   
\usepackage{xcolor} 

\usepackage{xcolor}
\usepackage[suppress]{color-edits} 
\addauthor{nn}{red}
\addauthor{zt}{olive}

\usepackage{libertinust1math}
\usepackage[nameinlink]{cleveref}

\usepackage{microtype}

\usepackage[commandnameprefix=ifneeded, todonotes={textsize=small}]{changes}
\definechangesauthor[name=ZT, color=olive]{z}
\usepackage{todonotes}
\renewcommand{\revision}[1]{#1}

\renewcommand{\final}[1]{#1}
\renewcommand{\finalZ}[1]{#1}

\title{FedMental: \final{Evaluating} Federated Learning \\for Mental Health Detection \final{from Social Media Data}}

\author{
  Nuredin Ali Abdelkadir$^{1}$, 
  Anjali Ratnam$^{1}$, 
  Zeerak Talat$^{2}$, 
  Stevie Chancellor$^{1}$ \\
  \vspace{1em} \\
  $^{1}$University of Minnesota \\
  \texttt{\{ali00530,ratna038,steviec\}@umn.edu} \\
  $^{2}$University of Edinburgh \\
  \texttt{ztalat@ed.ac.uk}
}

\usepackage{xspace}
\newcommand{\ie}{i.e.\@\xspace}

\begin{document}
\maketitle
\begin{abstract}

Social media text data are often used to train \final{Machine Learning} (ML) \final{models} to identify users exhibiting high-risk mental health behaviors. However, sharing this sensitive data poses privacy risks and limits the growth of benchmark datasets. We comprehensively evaluate whether privacy-preserving ML techniques can enable safer data sharing while preserving performance. Specifically, we apply federated learning (FL) and \nnedit{Differentially Private FL} for two widely-studied mental health prediction tasks: depression detection on X (Twitter) and suicide crisis detection on Reddit. We simulate realistic data-sharing scenarios by treating each user as a client in a non-IID setting, evaluating across different client fractions, aggregation strategies, and privacy budgets. While FL achieves comparable performance to centralized training
\final{(centralized $F1=85.63$; best FL model $F1=83.16$) on depression identification}, \finalZ{we find that} Differentially Private FL \ztdelete{presents}\finalZ{has} a large performance-privacy trade-off (\final{up to} $F1=27.01$ drop) even with low levels of noise ($\epsilon$ = $50$). \revision{This is due to the distortion of highly informative yet sparse mental health linguistic markers related to mental health, like health topics and emotion words.} This research empirically demonstrates the potential and limitations of current privacy preservation techniques for mental health inference tasks.

\end{abstract}

\section{Introduction}
\label{sec:introduction}
   
\ztedit{In the last decade, social media has emerged as an important source of data for predicting mental health conditions~\cite{de2014mental, yang2024mentallama} such as early detection of many mental illnesses~\cite{coppersmith-etal-2016-exploratory,benton2017multitask}, inference about support in online communities~\cite{pendse2019cross}, and crisis interventions~\cite{teague2022methods}. 
Approaches to these tasks have explored the use of data across modalities, including text, visual data, and interaction/engagement metadata~\cite{10.1145/3372278.3391932, 7885098}. 
Stigma and risk management around mental health disclosures inhibit honest disclosure online, and data can be difficult to source~\cite{chancellor2023contextual, fiesler2018participant}. 
Yet, data which contains personal information, particularly as it pertains to healthcare is protected data, and is regulated in the European Union and United Kingdom~\cite{GDPR2016a,WhatSpecialCategory2025}. 
Such privacy concerns apply in research as well: a vast majority of users in public mental health datasets are not properly de-identified~\cite{ayers2018don}.
}
\ztdelete{Social media is a crucial source of data for mental health prediction tasks~\cite{de2014mental, yang2024mentallama}, where applications  include early detection of many mental illnesses~\cite{coppersmith-etal-2016-exploratory,benton2017multitask}, inference about support in online communities~\cite{pendse2019cross}, and crisis interventions~\cite{teague2022methods}. These tasks rely on data from different modalities such as text, visual data, and interaction/engagement metadata~\cite{10.1145/3372278.3391932, 7885098}, and have recently explored pre-trained models~\cite{ji2021mentalbert,ji2023domain} and LLMs~\cite{yang2024mentallama}. 
}

\ztdelete{However, data about mental health is sensitive, and data privacy is a major concern. Stigma and risk management around mental health disclosures inhibit honest disclosure online, and data can be difficult to source~\cite{chancellor2023contextual, fiesler2018participant}. 
Even if this data can be found publicly, it is legally classified as protected in the European Union and the United Kingdom~\cite{GDPR2016a,WhatSpecialCategory2025}. These privacy concerns are well-grounded: a vast majority of users in public mental health datasets are not properly de-identified~\cite{ayers2018don}.}

\ztedit{Data protection protocols like Ethics Review Boards and Data Use Agreements help protect user data, but shift responsibility onto data controllers, \final{and} present institutional barriers to researchers who do not have access to institutional review~\cite{zirikly2019clpsych, shing-etal-2018-expert, coppersmith2015clpsych}.\ This also makes lapses in enforcement risks, exposing sensitive user data.\ Thus, benchmark datasets in this area are hard to find and can be small~\cite{chancellor2020methods}, limiting the growth of benchmark datasets.} Yet methods for ensuring the privacy of users' data in this domain are vital but remain under-explored.

\ztdelete{While data protection protocols, like Ethics Review Boards and Data Use Agreements, do help protect user data, they assign responsibility to the data controllers and can present institutional barriers to researchers who do not have access to institutional review~\cite{zirikly2019clpsych, shing-etal-2018-expert, coppersmith2015clpsych}. Moreover, lapses in enforcing agreements risk exposing sensitive user data. Collectively, this calls for considering technical methods for ensuring the privacy of users' data. To the best of our knowledge, comprehensive evaluations of the performance of computational strategies to ensure data privacy in this domain remain understudied.}

\ztdelete{
There is therefore a need for considering data privacy protection methods beyond . 
The primary methods for ensuring privacy is through Ethical Review Board approvals, exemption letters, and/or Data Usage Agreements~\cite{zirikly2019clpsych, shing-etal-2018-expert, coppersmith2015clpsych}. 
There are very few privacy mitigation strategies taken in this domain, and this is well-documented as a critical gap~\cite{ajmani2024data}. 
Many datasets are publicly available, with limited privacy preservation (e.g., usernames removed). 
If an approach is used, the primary approach for privacy protection is through 
Ethics Review Board approval, exemption letters, and/or Data Usage Agreements~\cite{zirikly2019clpsych, shing-etal-2018-expert, coppersmith2015clpsych}. 
While data protection protocols do help protect user data, they assign responsibility to the data controllers and can present institutional barriers to researchers who do not have access to institutional review. 
Furthermore, lapses in the enforcement of data protection agreements risk exposing sensitive user data. 
Comprehensive evaluations of the performance of computational strategies to ensure data privacy in this domain are, to the best of our knowledge, understudied. }

In this work, we comprehensively evaluate two data privacy strategies for the most common tasks in mental health inference in English:\ 
depression detection on X \finalZ{(Twitter)} and suicidal thought and behavior (STB) detection on Reddit.\ First, we evaluate Federated Learning (FL), where a global model is trained by distributing computation onto client devices, which train the model locally and send parameter updates, rather than raw data (i.e., text) to the global model. However, while FL can provide additional layers of privacy compared to centralized learning, it does not provide 
{\it guarantees} of privacy. 
Thus, we evaluate the combination of FL with differential privacy (DP) to study the utility-privacy tradeoffs. 
DP adds calibrated noise to limit inferences about individual users.

\ztdelete{Our results show}
\ztedit{We find} that the Federated setting is promising for these prediction tasks in both performance loss and efficiency; \revision{however, adding DP to FL creates an untenable utility-privacy tradeoff due to distortions of important yet sparse linguistic features.} For FL, there are no statistically significant differences in performance between FL and centralized models; FL models also require less data to obtain comparable performance. 
However, in comparison to the federated setting, the DP-FL models result in a large performance-privacy trade-off (F1=$27.01$ drop) even with low levels of noise ($\epsilon$ = $50$). 

Our results suggest that FL is a viable option; however, mathematically ensuring privacy \ztedit{using DP}\ztdelete{by adding noise} may not be practical due to significant performance drops and the negative impact of errors in mental health inference.

\section{Related Work}
\label{sec:related_work}

\paragraph{Mental Health Prediction.} 
\ztedit{ML has been applied to mental health prediction} from social media \ztedit{to identify}high-risk mental health behaviors and disorders.\ ML techniques have been used to predict behaviors and disorders at a post or user level \cite{alsagri2020machine, aldarwish2017predicting, islam2018depression, govindasamy2021depression}. 
\final{These methods rely on data across different modalities such as text, visual data, and interaction/engagement metadata~\cite{10.1145/3372278.3391932, 7885098}, and have recently explored language models pre-trained on mental health corpus~\cite{ji2021mentalbert,ji2023domain}. These works rely on a centralized training approach, where users' mental health disclosure data is shared with researchers or scraped and labeled from social media.}

\paragraph{Federated Learning.} 
Federated learning is a decentralized training mechanism for machine learning where \ztedit{client (edge) devices collaboratively train a shared global model~\cite{li2020review,mcmahan2016federated} while avoiding the transfer of raw data (e.g., text)}. Within the FL training paradigm, clients contribute to training the global model by first receiving a copy of the model, which is then trained on locally on the client device on locally held data.\ 
Finally, clients share the local model updates with the global server~\cite{mcmahan2016federated}.\
The collected updates are then aggregated using algorithms such as Federated Averaging~\cite[FedAvg,][]{mcmahan2017communication}, FedProx~\cite{li2020federated}\ztdelete{, which adds a proximal term to FedAvg}, FedOPT~\cite{reddi2020adaptive}\ztdelete{\final{, etc}}\ztdelete{, which \ztdelete{introduces}\ztedit{maintains an optimizer on the global server that is distinct from the optimizers on the client devices.}}\ztdelete{ a separate optimizer on the global server.}. 

\revision{Federated learning has been applied in mental health domains with similar risks from different data sources.\ \citet{khalil2024exploring} surveyed 16 papers on federated learning within psychiatric tasks, the most common being depression detection, specifically used to 
predict durations of hospitalization using Electronic Health Records~\cite{pfohl2019federated}, 
detect depression using wearable sensors~\cite{aminifarMonitoringMotorActivity2021, wang2024differential, gupta2024privacy}, or self-reported assessment data sources \cite{kuang2025federated}.
Others used mobile device data (keystrokes and accelerometer values) and clinical surveys to predict mood, and found that training on the IID setting yields better performance ~\cite{xu2021fedmood}.}

Using social media as a data source, federated approaches to mental health detection have also been explored~\cite{vasconcelos2023exploring,basu2021benchmarking,liu2024depression,Ji_Detecting_2019}.\ 
However, prior work has either focused on either IID settings or classifying individual posts.\
For instance, \citet{vasconcelos2023exploring} applied a federated approach to the eRisk depression dataset to predict whether a post is labeled as depression or control.
\citet{basu2021benchmarking} explored federated settings in both IID and non-IID data distribution using BERT-based models to predict depression and sexual harassment posts. \citet{liu2024depression} uses federated learning on cross-platform and multilingual social media data to predict depression. 
Finally, \citet{Ji_Detecting_2019} trains a federated CNN \final{and LSTM classifiers and proposes an advanced optimization scheme for data protection learning framework (AvgDiffLDP)} for predicting suicidal ideation on Reddit in an IID setting. 
Data privacy concerns are underexplored in these domains in a few ways. 
First, these works \final{lack} evaluating data privacy considerations holistically (e.g., modeling the tradeoffs between privacy preservation and performance \final{across various client fractions and aggregation algorithms}). 
Second, most of these approaches do not use a naturalistic non-IID setting of FL \final{representing each user with their post history as a separate client}.

\paragraph{Differential Privacy.}
\ztedit{While federated learning affords an additional layer of data protection compared to centralized training, which requires sharing raw data, differential privacy is a mechanism for obtaining privacy through the addition of calibrated noise to the training process~\cite{Shan2024}. 
The addition of the calibrated noise seeks to mitigate the identification of any individual data point while maintaining global aggregates and patterns, which can be used for training models under mathematical guarantees of privacy. 
Thus, where federated learning affords privacy of data by not sharing raw data, differential privacy maintains privacy by adding noise to each data point, and can be used independently of federated settings. 
}

Prior work at the intersection of federated learning and differential privacy has investigated different aspects of machine learning for mental health. 
\citet{basu2021benchmarking} have examined the identification of sexual harassment and depression in individual posts; they found that utility degradation is higher in a non-IID than an IID setting, and noise addition has more effect when training on a small dataset size.
Recently, \citet{sarwar2025fedmentor} applied DP fine-tuning exclusively to Low-Rank Adaptation~\cite{hu_lora_2021} to reduce communication and memory consumption.

\section{Experimental Setup}
\label{sec:experimental_setup}

\subsection{Datasets}

We 
selected benchmark datasets for high-risk behaviors and disorders, specifically depression and suicidality, from the most popular tasks in MH inference, from X and Reddit. We focus on user-level prediction from users' historical posts from five relevant datasets.\ These disorders, datasets, and platforms have been 
examined in prior work~
\cite{chancellor2020methods}.

\begin{table}[h!]
\centering 
    \small 
    \resizebox{\linewidth}{!} {      
\begin{tabular}{@{}llccc@{}}
        \toprule
        \textbf{Disorder} & \textbf{Classes} & \textbf{Train} & \textbf{Validation} & \textbf{Test} \\
        \midrule
        \multirow{2}{*}{Depression} & Treatment & 1,844 & 263 & 528 \\
        & Control & 2,123 & 303 & 608 \\
        \midrule
        \multirow{2}{*}{Suicide} & Treatment & 597 & 85 & 171 \\
        & Control & 534 & 76 & 153 \\
        \bottomrule
    \end{tabular}                      
}
\caption{Datasets used in our experiments.}
\label{tab:datasets}    
\end{table}

\subsubsection{Depression}

We used three datasets from X, as about 50\% of datasets for detecting depression rely on X~\cite{aldkheel2024depression}. 
These datasets provide user post histories that enable user-level predictions and are collected using keywords and phrases that disclose depression, such as \textit{``(I'm / I was / I am / I've been) diagnosed with depression.''}

\paragraph{CLPsych.} CLPsych was 
released with the 
2015 CLPsych Shared Task for Depression~\cite{coppersmith2015clpsych}.\
The data is split into treatment (\ie, users who self-disclose diagnoses) and a control group. It is manually annotated to verify the authenticity of the disclosures.\ 
The dataset contains up to 3,000 posts for each user, with the self-disclosure posts removed. 
We use 477 users labeled as Depression, and the 871 control users who do not disclose depression.

\paragraph{MTL-D.} The Multitask Learning-Depression (MTL-D) dataset~\cite{shen2017depression} contains one month of user post history for 1,840 users labeled as depressed and 1,840 labeled as control users. The posts in the dataset contain images and textual data. In our work, we make use of the textual data.

\paragraph{
CCD.} The Cross-Cultural Depression (CCD) dataset~\cite{abdelkadir2024diverse} contains up to 3,200 posts for 267 users labeled as treatment (\ie, depressed) and 264 labeled as control users. This dataset is manually annotated and samples from seven English-speaking countries that are culturally and geographically diverse.

\subsubsection{Suicidal Thoughts and Behaviors (STB)} 
Reddit has emerged as the predominant data source for ML for identifying suicidal thoughts and behaviors. There are dedicated mental health-support subreddits about this topic, such as r/SuicideWatch, r/selfharm, and r/StopSelfHarm. 
These subreddits can offer insight into the language of users who display STB and of those who present a high risk of engaging in suicidal behaviors. 

Following prior work~\cite{chancellor2020methods, shing2018expert}, we split users into treatment (at-risk users) and control groups (very low or no risk users).

\paragraph{C-SSRS.} The Columbia Suicide Severity Rating Scale dataset (C-SSRS) dataset~\cite{gaur2019knowledge} was constructed for predicting suicide risk by categorizing users from mental health fora on Reddit into five groups following the C-SSRS, ranging from least to most concerning: supportive (110 users), indicator (100 users), ideation (170 users), behavior (75 users), and attempt (45 users). 
In our work, we group users labeled as ideation, behavior, and attempt as our treatment group of STB (290 users), and supportive and indicator (210 users) as our control group.

\paragraph{UMD-RD.} The UMD Reddit Suicidality dataset (UMD-RD) dataset~\cite{shing-etal-2018-expert, shing2020prioritization}  
is manually labeled to verify that they exhibit genuine STB using four levels of suicide risk: No Risk, Low Risk, Moderate Risk, and Severe Risk. 
The control group consists of users who 
did not post in any mental health-related subreddits while the treatment group are sampled from users who have posted to the \textit{r/SuicideWatch} subreddit. 
We group users with Moderate (256) and Severe (302) Risk into our treatment group (358 users), and users labeled as No Risk (195) and Low Risk (113) into our control group (308 users).

\subsection{Preprocessing} 

We combine the three datasets for identifying depression and the two datasets for identifying STB into a depression dataset and an STB dataset, respectively.\
We then perform a stratified split of the combined datasets into training (70\%), validation (10\%), and test (20\%) sets.\footnote{We include validation and test splits into the training set, for datasets in which they are separated.}
See \Cref{tab:datasets} for summaries of the dataset splits.\
We then preprocess the datasets to remove retweet tokens, username mentions, URLs, and numeric values, and expand English word contractions.

\subsection{Models}


We conducted our experiments with one linear and six transformer-based pre-trained models. Our linear model is a Logistic Regression, chosen for its quick training time, competitive performance, and interpretable predictions~\cite{benton2017multitask, jiang2018detecting, harrigian2020models}.\ 
For the transformer-based models, we use the general-purpose architectures BERT~\cite{kenton2019bert} and RoBERTa~\cite{liu2019roberta}, and their distilled counterparts: DistilBert and DistilRoBERTa~\cite{sanh2019distilbert}, as these models have been used for classification tasks~\cite{aftan2023survey}. 
Following recent work, we also experiment with two mental health-specific models: MentalBERT~\cite{ji2021mentalbert}, which has been trained on $\sim$13 million sentences from Reddit subreddits for discussing depression, anxiety, and suicide topics; and MentalLongformer~\cite{ji2023domain}, which is optimized for longer token sequences.. 

For all transformer-based models, we replace the pre-trained head of the transformer models with a randomly initialized classification head prior to conducting the training. 
We train the Logistic Regression model and fine-tune the transformer-based models to classify users as depression or control, and STB or control, respectively, based on their post history. 
\ztedit{We do not experiment with zero-shot settings for two reasons. First, a zero-shot setting requires transmitting data to a centralized server, thereby foregoing preservation of privacy. Second, prior work reports sub-par performance in zero-shot settings for the mental health domain 
\cite[see e.g,][]{yang-etal-2023-towards}.
}
\ztdelete{\nnedit{Zero-shot settings have been found to result in subpar performance in this domain \cite{yang-etal-2023-towards}, and privacy concerns may arise since data still needs to be transferred to the model owners. Therefore, we do not opt for zero-shot settings.}}

\subsection{Training Schema}
Next, we discuss the centralized and federated learning approach applied to our experiments, applied to the seven models above.

\subsubsection{Standardized/Centralized Approach} 
The centralized setting serves as a baseline. 
This standard ML approach requires datasets and models to be located in the same location, or on a single server. For all transformer-based models, we replace the pre-trained heads with randomly initialized classification heads. We train our models for 50 epochs and set early stopping to 5 epochs. Please refer to \Cref{app:sec:replication} for further details on model setup and hyperparameters.

\subsubsection{Federated Learning Approach}

For FL, we conceptualize client devices as individual users in the datasets. Our setup implies that the labels are not uniformly distributed across clients. However, the number of clients in the control and treatment groups is balanced. 
Moreover, the user histories that are available vary across clients (see depression (\Cref{fig_dep_treatment_tokens,fig_dep_control_tokens}) and STB (\Cref{fig_suicide_treatment_tokens,fig_suicide_control_tokens}) in \Cref{sec:token_dist}). 
Thus, our models are not trained on independent and identically distributed (IID) data. 
Prior work \cite[e.g.,][]{gandhi_federated_2022,xu2021fedmood} has found that IID setups result in higher performances than non-IID setups for federated learning.\ This affords a realistic use case, in which a person accessing healthcare or using social media for well-being enrolls in a monitoring system, or a user consents to share their data with a research team.

\paragraph{Training.}
We train our models for 100 rounds on the server and train clients for 50 epochs per round. 
That is, we perform 100 rounds of retrieving client updates and aggregate them on the server, and we train each client model for 50 epochs in each round.\ 
We experiment with four different client fractions (c = $10, 30, 50,$ and $70$), which represent the percentage of client devices/users included in the aggregation step on the server in each round (i.e., at $c=10$, 10\% of all clients are randomly sampled for the inclusion of their updates in the global model).\footnote{Selecting clients for inclusion in the aggregation step on the server is an open research area with different benefits and drawbacks to each aggregation method~\cite{fu2023client,gouissem2024comprehensive}.}  
We set the client learning rate to $4e-5$ and the server learning rate to $1e-3$. \ztedit{See \Cref{app:sec:replication} for full experimental details.} 
\ztdelete{Please refer to \Cref{app:sec:replication} for further details required for replication.}

We experiment with three different aggregation algorithms: FedAvg, FedProx, and FedOPT. 
FedAvg is a common aggregation algorithm and serves as our baseline federated method, taking the weighted average of the received client updates. 
FedProx and FedOPT address issues of slow convergence with the FedAvg algorithm~\cite{moshawrab2023reviewing}. 
FedProx addresses the poor suitability of FedAvg for heterogeneous data situations by adding a proximal term, while FedOPT introduces separate optimizers for client and server models to introduce adaptive optimizers to federated learning. 

In our experiments, we set the proximal constant $\mu=0.01$ for FedProx and apply a server optimizer for FedOPT. 
All other hyperparameters are shared across the three algorithms.

\subsection{Differentially Private FL}
\label{dpfl_method}
\ztedit{
For our differentially private FL (DP-FL) setup, we implement a client-level DP mechanism. Each client trains a model locally and computes the model update in the traditional federated learning setup. 
Prior to transmitting the update vector from client to server, we apply an $\ell_2$-norm clipping to the update vector and add calibrated Gaussian noise. This is adaptively scaled as a factor of the training round.\ 
Next, on the server, we evaluate the resulting utility-privacy trade-off across different privacy budgets ($\epsilon, \delta$=$1e-5$), where the total accumulated  privacy loss is tracked in each round (see \Cref{app:sec:replication} for further experimental details).\ 
\nnedit{We \ztdelete{utilized}\ztedit{used} the best-performing models and hyperparameters from the traditional FL setting as our baseline. 
Specifically, we fine-tune a BERT model for the depression identification task using the FedProx aggregation methods.\ \final{To investigate whether a change in the pre-trained model, aggregation algorithm, or the inference task affects the differentially private FL, we conduct additional experiments utilizing MenatlBERT as a model, FedAvg aggregation algorithm across both prediction tasks (depression and suicidal thoughts and behaviors)}. The primary goal of this investigation is to provide a comprehensive feasibility study of DP-FL under these conditions.}
}

\final{To ensure the reliability of our findings, we conducted stability analyses for representative models across all settings. We report 95\% Confidence Intervals (CIs) calculated over five random seeds for these representative configurations in the centralized, standard FL, and DP-FL settings.}

\section{Results}
\label{sec:results}

We evaluate the performance of our models using the F1 score and recall.
The best-performing model for both depression detection ($F1=85.63$, see \Cref{tab: depression}) and suicidal ideation ($F1=85.44$, see \Cref{tab: suicide}) is the centralized MentalLongformer model, which has been pretrained on mental health subreddits. \final{In the standard federated setting (without differential privacy), models achieved competitive performance: MentalLongformer using only 10\% of client fractions reached $F1=83.16$ for depression (\Cref{tab: depression}), and MentalBERT using 50\% of client fractions reached $F1=84.09$ for suicidal ideation (\Cref{tab: suicide})}. This suggests that beyond privacy preservation, federated learning offers a computationally efficient alternative to centralized training. \final{We present detailed findings for depression identification (\Cref{depression_analysis}), suicidal thoughts and behaviors (\Cref{stb_analysis}), data efficiency (\Cref{data_efficiency_analysis}), the effect of differential privacy (\Cref{dpfl_findings}), and an analysis of the resulting privacy-utility trade-off due to differential privacy (\Cref{utility_privacy_tradeoff_discussion}).}

\subsection{Depression Analysis}
\label{depression_analysis}

Federated approaches perform similarly to the centralized training approach for depression detection (see \Cref{tab: depression}). 
While there is minimal drop in performance for federated models, the decrease is not statistically significant ($p=0.21875$, $a=0.05$ using the Wilcoxon signed-rank test).\ 
The largest performance difference between federated and centralized models is $4.61$ point drop in F1 score by the Logistic Regression classifier. 
For transformer models, the largest performance drop of federated models is $2.47$ point drop in F1 score.\ 
There are also models for which the federated settings outperformed the centralized setting. 
Specifically for MentalBERT, DistilBERT, and DistilRoBERTa we see small performance gains between $0.26$ and $0.66$ points in F1-score. \revision{\ztedit{See \Cref{tab:fedavg_depression_results,tab:fedprox_depression_results,tab:fedopt_depression_results} in \Cref{app:sec:dep_suicidal_results} for full results}}\ztdelete{\revision{See Table \ref{tab:fedavg_depression_results} for fedavg, Table \ref{tab:fedprox_depression_results} -- fedprox and Table \ref{tab:fedopt_depression_results} -- fedopt full results in the  \Cref{app:sec:dep_suicidal_results}}}.



\begin{table}[h!]
\centering
    \small
    \resizebox{
    \columnwidth}{!} {
\begin{tabular}{@{}l*{4}{c}@{}}
        \toprule
        \multirow{2}{*}{\textbf{Model}} & \multicolumn{2}{c}{\textbf{Centralized}} & \multicolumn{2}{c}{\textbf{Federated}} \\
        \cmidrule(lr){2-3} \cmidrule(l){4-5}
        & Recall & F1 & Recall & F1 \\
        \midrule
        Logistic Regression & \textbf{73.59} & \textbf{73.67} & 69.03 & 69.06 \\
        \addlinespace
        MentalBERT & 76.94 & 76.74 & \textbf{77.25} & \textbf{77.08} \\
        \addlinespace
        MentalLongformer & \textbf{85.82} & \textbf{85.63} & 82.99 & 83.16 \\
        \addlinespace
        BERT & \textbf{78.73} & \textbf{78.86} & 77.62 & 77.57 \\
        \addlinespace
        RoBERTa & \textbf{79.26} & \textbf{79.03} & 76.85 & 76.83 \\
        \addlinespace
        DistilBERT & 76.71 & 76.76 & \textbf{77.33} & \textbf{77.42} \\
        \addlinespace
        DistilRoBERTa & 79.32 & 79.47 & \textbf{79.63} & \textbf{79.73} \\
        \bottomrule
    \end{tabular}
}
\caption{Depression models comparing the centralized approach vs the best \ztdelete{performance achieved by the}\ztedit{performing} federated setting.}
    \label{tab: depression}
    \vspace{-9pt}
\end{table}

\subsection{STB Analysis} 
\label{stb_analysis}

For identifying users with STB, federated models sometimes outperform their centralized counterparts (see \Cref{tab: suicide}).\ 
For instance, the federated Logistic Regression model ($5.50$ points in F1 score) and federated MentalBERT ($4.12$ points in F1 score) outperform their centralized counterparts. 
Indeed, for three models for identifying STB, models trained using the federated approach outperform the centralized models.\ 
While this is encouraging for federated models for identifying 
STB, our analysis indicates no statistically significant differences between performances ($p=0.6875$, $a=0.05$ via Wilcoxon signed-rank test).\ \revision{\ztedit{See full results in \Cref{tab:fedavg_suicide_results,tab:fedprox_suicide_results,tab:fedopt_suicide_results} in \Cref{app:sec:dep_suicidal_results}. }}


\begin{table}[htbp]
    \centering
    \label{tab:model-comparison}
    \small
    \begin{tabular}{@{}l*{4}{c}@{}}
        \toprule
        \multirow{2}{*}{\textbf{Model}} & \multicolumn{2}{c}{\textbf{Centralized}} & \multicolumn{2}{c}{\textbf{Federated}} \\
        \cmidrule(lr){2-3} \cmidrule(l){4-5}
        & Recall & F1 & Recall & F1 \\
        \midrule
        Logistic Regression & 60.50 & 59.62 & \textbf{65.24} & \textbf{65.12} \\
        \addlinespace
        MentalBERT & 79.84 & 79.97 & \textbf{83.95} & \textbf{84.09} \\
        \addlinespace
        MentalLongformer & \textbf{85.43} & \textbf{85.44} & 81.56 & 81.73 \\
        \addlinespace
        BERT & 81.48 & 81.60 & \textbf{82.65} & \textbf{82.80} \\
        \addlinespace
        RoBERTa & \textbf{82.59} & \textbf{82.63} & 81.51 & 81.62 \\
        \addlinespace
        DistilBERT & \textbf{79.22} & \textbf{79.35} & 77.81 & 77.75 \\
        \addlinespace
        DistilRoBERTa & 81.15 & 81.27 & \textbf{81.63} & \textbf{81.79} \\
        \bottomrule
    \end{tabular}
    \caption{Suicide models comparing the centralized approach vs the best \ztdelete{performance achieved by the}\ztedit{performing} federated setting.}
     \label{tab: suicide}
\end{table}

\subsection{Efficiency Analysis for FL}
\label{data_efficiency_analysis}

We further find that in some instances, federated models compete with centralized models while using only $10-50$\% of the total available data. Comparing the different aggregation algorithms trained on different client fractions with the centralized approach, we observe that training on $10\%$ offers the best performance. For instance, see MentalBERT and DistilRoBERTa trained only on $10\%$ of the data outperforms the centralized approach and other trainings on larger client fractions in the depression and STBs identification cases (see Table \ref{tab:fedprox_depression_results} and \ref{tab:fedprox_suicide_results} \revision{in \Cref{app:sec:dep_suicidal_results}}). Increasing the client fraction does not correspond to a significant performance increase in either F1 or recall.

Similar to \newcite{galaFederatedApproachHate2023}, we argue that data efficiency gains may be due to the selected clients being a representative sample of the overall data. 
Client selection can, therefore, constitute an interesting area for future work. 


\subsection{Differentially Private FL}
\label{dpfl_findings}
\begin{figure}[ht]
  \centering
  \includegraphics[width=0.5\textwidth]{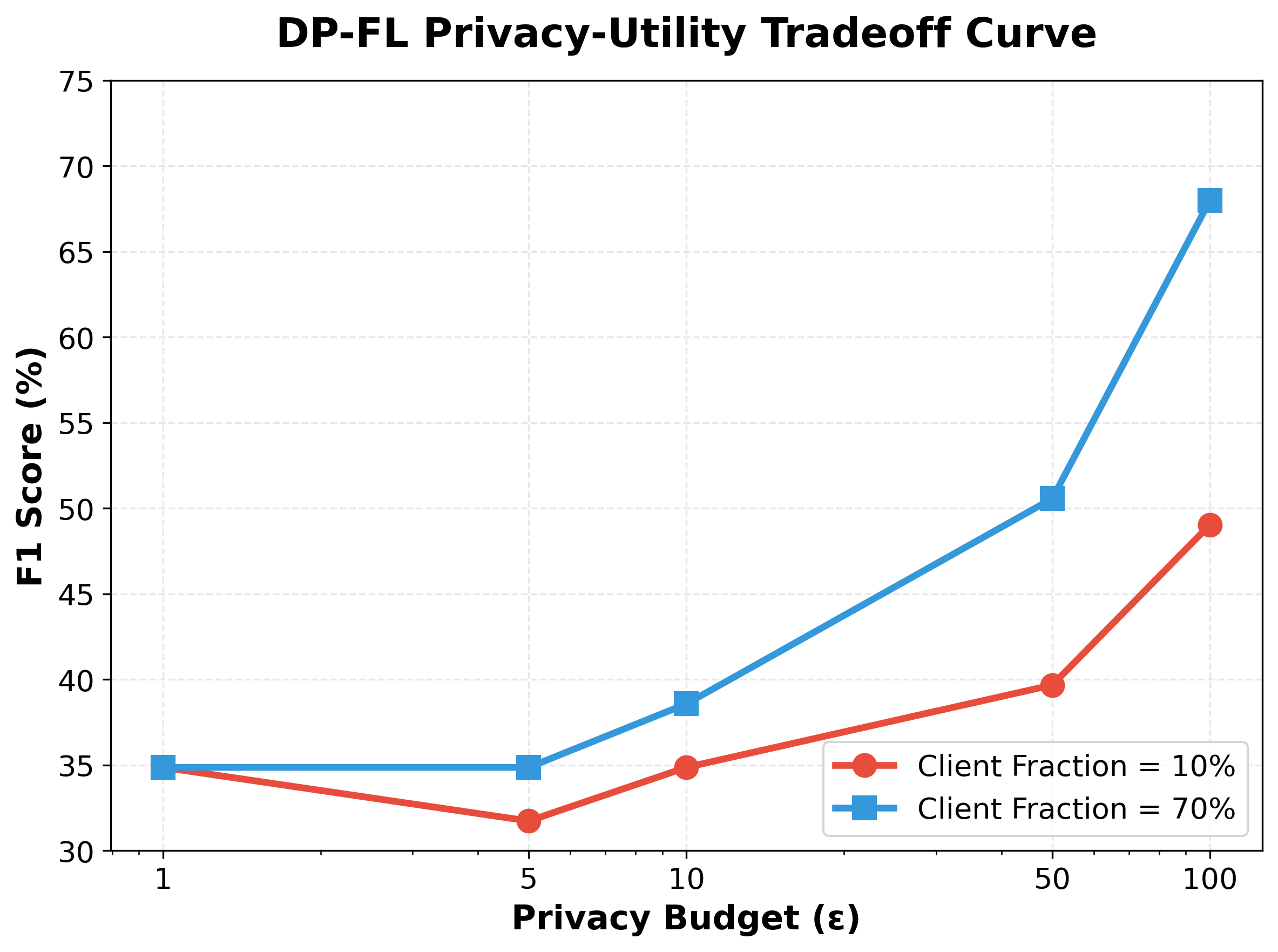}
 \caption{\revision{Utility-privacy trade-off \ztdelete{among}\ztedit{for} different client fractions and privacy budgets for depression detection. \ztdelete{An increased privacy leads to lower utility. The lower client fraction also contributes to reduced performance in this setting}}}
 \label{fig_dep_utility_privacy_trade_off}
\end{figure}

\ztedit{Our experiments with the DP-FL setting show a large utility drop of $27.01$ F1 ($F1_{FL}=77.57\rightarrow F1_{FL-DP}=50.56$) for depression identification, where $\epsilon$=$50$ and $c$=$0.7$ (see \Cref{tab:DP_DPFL_results} and \revision{\Cref{fig_dep_utility_privacy_trade_off}}). 
As the strength of the privacy increases with lower values of $\epsilon$, the performance further degrades. 
For a strong privacy budget ($\epsilon$ $= 5$), the drop is even more significant (F1 = $34.86$). 
In line with prior work, e.g., \citet{tramer2020differentially}, we find that smaller client fractions result in larger drops in utility, thereby negating data efficiency benefits of the pure FL setup.
Further, such large drops in utility challenge the viability of the DP-FL setting compared to centralized and federated settings. \revision{We found similar trends in the STB task (\Cref{tab:STB_DPFL_results}).}
}
\ztdelete{
\nnedit{The differentially private FL resulted in a large utility drop off ($28.54$ f1) [f1 = $77.57$ best performing FL to f1 = $50.56$ in DP-FL] in the depression identification task. This performance drops with even higher privacy budgets ($\epsilon$ $= 50$). A strong privacy budget ($\epsilon$ $= 5$) results in a much lower utility (f1 = $34.86$). This privacy utility drop-off represents a significant drop, potentially resulting in a classifier that makes DP-FL in this setting not a viable option compared to the centralized or FL settings no mathematical privacy guarantee. We also observe a reduced performance when applying noise to lower client fractions (c=$0.1$) compared to larger client fractions (c=$0.7$). This highlights the 
 Table \ref{tab:DP_combined_results} reports the DP-FL results.   
}
}

\begin{table}[htbp]
\centering
\small
\setlength{\tabcolsep}{3pt}

\begin{tabularx}{\columnwidth}{@{}lXXXX@{}}
\toprule
\textbf{Privacy Budget} & \multicolumn{2}{c}{\textbf{Client Frac. 10\%}} & \multicolumn{2}{c}{\textbf{Client Frac. 70\%}} \\ \cmidrule(lr){2-3} \cmidrule(lr){4-5}
($\epsilon$) & \textbf{Recall} & \textbf{F1} & \textbf{Recall} & \textbf{F1} \\ \midrule
1   & 50.00 & 34.86 & 50.00 & 34.86 \\
5   & 50.00 & 31.73 & 50.00 & 34.86 \\
10  & 50.00 & 34.86 & 49.12 & 38.57 \\
50  & 50.33 & 39.67 & 54.34 & 50.56 \\
100 & 49.27 & 49.03 & 67.83 & 67.97 \\ \bottomrule
\end{tabularx}

\caption{\revision{\ztedit{DP-FL performance across different privacy budgets and client fractions. Larger client fractions improve the DP utility-privacy trade-off.}}}
\label{tab:DP_DPFL_results}
\end{table}

\begin{table}[htbp]
\centering
\small
\setlength{\tabcolsep}{3pt}
\revision{
\begin{tabularx}{\columnwidth}{@{}lXXXX@{}}
\toprule
\textbf{Privacy Budget} & \multicolumn{2}{c}{\textbf{Client Frac. 10\%}} & \multicolumn{2}{c}{\textbf{Client Frac. 70\%}} \\ \cmidrule(lr){2-3} \cmidrule(lr){4-5}
($\epsilon$) & \textbf{Recall} & \textbf{F1} & \textbf{Recall} & \textbf{F1} \\ \midrule
1   & 50.00 & 34.55 & 50.00 & 32.07 \\
5   & 50.00 & 34.55 & 50.00 & 32.07 \\
10  & 45.35 & 32.08 & 50.00 & 30.91 \\
50  & 46.34 & 46.28 & 66.25 & 65.58 \\
100 & 47.11 & 34.35 & 70.02 & 69.07 \\ \bottomrule
\end{tabularx}
}
\caption{\revision{\ztdelete{Suicidal Thoughts and Behaviors -- Comparison of DP-FL performance across different privacy budgets for 10\% and 70\% client fractions.}\ztedit{Suicidal Thoughts and Behaviors--DP-FL performance across privacy budgets and client fractions.}}}
\label{tab:STB_DPFL_results}
\end{table}

\final{To better understand whether this privacy-utility drop-off introduced when applying DP-FL is an artifact of the model, aggregation selection, or the task, we discuss our experimental findings on different settings on 95\% confidence intervals over 5 random seeds.
The results show a large utility-privacy trade-off regardless of model or aggregation choice.\ For instance, for depression identification, MentalBERT with FedAvg achieves $F1 = 0.3361$ ± $0.0207$ ($\epsilon = 10$, $c = 0.1$) and $0.4599$ ± $0.0287$ ($\epsilon = 100$, $c = 0.1$), comparable to BERT with FedOpt $F1 = 0.3361$ ± $0.0172$ ($\epsilon = 10$, $c = 0.1$), despite changing both the model architecture and aggregation algorithm.\ Similarly, for STB identification, MentalBERT achieves $0.3798$ ± $0.0628$ ($\epsilon = 10$, $c = 0.1$) and $0.3915$ ± $0.0215$ ($\epsilon = 100$, $c = 100$). Both results remain far below standard FL ($F1 = 70.96$ STB; $76.85$ depression).\ Those findings show the trade-off is not an artifact of the model, aggregation selection, or task.}

\nnedit{
\final{Performance drops are generally expected when striving to achieve mathematical guarantees for privacy.}
Better convergence performance leads to \ztdelete{lower protection levels}\ztedit{lower levels of privacy protection}~\cite{wei2020federated}. 
\ztedit{For example, \citet{wei2020federated} find that there can be up to a $20$\% accuracy drop when privacy guarantees are increased.}
\ztdelete{There can be up to a 20\% accuracy drop when increasing privacy guarantees \cite{wei2020federated}.}
This significant drop in performance can be associated with the limited data, \ztedit{which is a common issue in the social media mental health detection domain. 
In settings with limited data, low levels of noise can distort the model training in DP settings~\cite{tramer2020differentially, jana2021investigation}. 
Similarly, although our non-IID setup is more realistic, it may also be a factor in the utility drop for DP-FL~\cite{basu2021benchmarking}. }
\ztdelete{as is the case in the social media mental health detection domain, which struggles with limited data, where little noise distorts the model training in DP settings \cite{tramer2020differentially, jana2021investigation}. Non-IID setup, which depicts a realistic scenario, as represented in our setting, is also a factor for a utility drop in a DP-FL \cite{basu2021benchmarking}.}   
}


\subsection{\final{Stability and Variance Analysis}}

\final{To assess the reliability of our results, we conducted stability analyses across five random seeds for representative models in each experimental setting (see \Cref{tab:stability}).\ From the sample of experiments conducted, we find that variance remains low in the centralized and standard FL settings (e.g., $SD \pm 0.0102$ for MentalBERT or $SD \pm 0.0298$ for DistilBERT). These variance intervals support the reliability of our single-run centralized and standard federated setting results.\ However, in the DP-FL setting, we observe increased instability ($SD \pm 0.0758$ for BERT at $c=0.7$), highlighting the trade-off between privacy guarantees and model training stability in such sensitive mental health inference tasks from social media.}
\begin{table}[ht]
\centering
\setlength{\tabcolsep}{3pt} 
\resizebox{\columnwidth}{!}{
\begin{tabular}{@{}lllcc@{}}
\toprule
\textbf{Setting} & \textbf{Model} & \textbf{Task} & \textbf{F1 ($\mu \pm \sigma$)} \\ \midrule
\textit{Centralized} & MentalBERT & STB & $0.8271 \pm 0.0102$ \\
\textit{Centralized} & BERT & STB & $0.8113 \pm 0.0190$ \\ \midrule
\textit{Standard FL} & DistilBERT\textsubscript{c=0.7} & Depression & $0.7617 \pm 0.0298$ \\
\textit{Standard FL}\textsuperscript{$\ast$}  & BERT\textsubscript{c=0.7} & STB & $0.6703 \pm 0.0420$ \\
\midrule
\textit{DP-FL}\textsuperscript{$\dagger$} & BERT\textsubscript{c=0.1} & STB & $0.3310 \pm 0.0140$ \\
\textit{DP-FL}\textsuperscript{$\dagger$} & BERT\textsubscript{c=0.7} & STB & $0.4134 \pm 0.0758$ \\

\textit{DP-FL}\textsuperscript{$\ddagger$} & BERT\textsubscript{c=0.1} & STB & $0.4246 \pm 0.0641$ \\
\textit{DP-FL}\textsuperscript{$\ddagger$} & BERT\textsubscript{c=0.7} & STB & $0.6849 \pm 0.0429$ \\\bottomrule
\end{tabular}%
}
\caption{Stability analysis across five random seeds. $c$ denotes client fraction. \textsuperscript{*}FedAvg. FedOPT aggregation is used for the rest.   \textsuperscript{$\dagger$}$\epsilon=10$, \textsuperscript{$\ddagger$}$\epsilon=100$. $\mu \pm \sigma$ denotes Mean $\pm$ SD. Standard and Centralized settings show some stability, while DP-FL introduces higher variance.}
\label{tab:stability}
\end{table}

\section{\revision{Privacy-Utility Trade-off}}
\label{utility_privacy_tradeoff_discussion}
\revision{
\ztedit{To understand the\ztdelete{se} performance drops in the DP-FL setting, we conduct two mixed-method analyses for depression detection.}
}

\subsection{\revision{Stronger Privacy Distorts Sparse Depression Linguistic Markers}}
\revision{
\ztedit{In the context of DP-FL, model performance deteriorates privacy guarantees increase, i.e., by decreasing values of $\epsilon$ (see \Cref{tab:DP_DPFL_results,tab:STB_DPFL_results}). 
While we expect \ztdelete{a} performance drop\finalZ{s} when applying privacy-preserving mechanisms, the performance drops we observe \finalZ{with DP-FL} are larger than expected. 
One \finalZ{potential} explanation is that the noise distorts important linguistic markers of mental illness. }
\ztdelete{Our findings show that within the DP-FL, the models perform worse as we increase the privacy guarantees by increasing the level of noise (Table \ref{tab:DP_DPFL_results} and \ref{tab:STB_DPFL_results}). This is similar to other tasks but much more pronounced for ours than is expected. Our hypothesis here is that, as we lower the privacy budget (epsilon) and increase the level of noise, the model distorts the sparse and important linguistic markers for identifying depression.} 
}
    
\paragraph{\textbf{\revision{Method.}}}
    
\revision{To evaluate why performance is dropping substantially, we compare three models trained at different privacy budgets while controlling all other hyperparameters.\ We compare ($\epsilon$ = 10, 50, 100), where a higher epsilon \ztdelete{is lower in}\finalZ{indicates lower} privacy \finalZ{protections}. In the DP-FL setting, these three models vary in performance ($F1$=$38.57$, $F1$=$50.56$, $F1$=$67.97$), respectively. We extracted top features using SHAP (SHapley Additive exPlanations) and qualitatively analyzed important features. Given the importance of socio-linguistic cues in mental health prediction, we also conducted LIWC analysis \cite{pennebaker2015development} on the top 50 features to identify psychological and emotional markers learned at different privacy budgets.}  

\paragraph{Result.} We find that increased noise does distort important markers for depression and relevant linguistic cues related to mental health. The model with the weakest privacy guarantees ($\epsilon$=100, F1=67.97), which is the best performing DP model, correctly learns 16\% of the top 50 keywords as pertinent, half of which are health-related keywords (e.g., ``tumor'', ``prescribed'', ``aching'', ``intermittent'') and the other half related to negative emotions (e.g., ``witnessed'', ``vanish'').\
    \revision{The LIWC-based analysis further confirms this; we find health, social/work, and religion-related terms that are all potential indicators of depression.}
    
    \revision{In contrast, with stronger privacy guarantees ($\epsilon=50$, $F1=50.56$), features learned by the model are irrelevant terms and patterns.\ For example, 22\% are generic keywords (e.g., ``premise'', ``traditionally'', ``confluence'') and 8\% are entertainment-related (e.g., ``Eurovision'', ``maverick'') keywords. 
    \finalZ{In our LIWC analysis, we find that there is minimal inclusion of keywords related to affect, negative emotion, or keywords related to health.}\ztdelete{Here, our LIWC analysis found minimal inclusion of affect, negative emotion, or health indicating keywords.}}

    \revision{
   The model trained with the strongest privacy guarantees ($\epsilon=10$, $F1=38.57$) identifies keywords related to random topics (e.g., ``Egypt'', ``penguins'', ``witchcraft'') \finalZ{which is supported by our LIWC-based analysis, in which we find the occurrence of}\ztdelete{while our LIWC-based analysis revealed} LWIC categories such as percept/see (e.g., term: graphics), affect (e.g., term: magnificent), drives/power/work (e.g., term: governments) -- which may not be direct indicators of depression.
    Across our three settings, we thus find a clear trend: the stronger the privacy guarantees and the stronger the noise distortion, the more models rely on irrelevant words for prediction. \finalZ{This supports our hypothesis that the performance drops in DP-FL may be caused by models not learning important mental health-related keywords due to noise disrupting the sparse linguistic signals that are critical for this task.} \ztdelete{thereby accounting for performance drops as important linguistic markers,} 
    }

\subsection{\revision{Smaller Client Fractions Impacted with Low Noise}}

\revision{Our findings indicate that smaller client fractions reduce the DP-FL performance, the opposite of standard FL, which improved data efficiency when trained on lower client fractions.
\ztedit{We therefore analyze the impact of noise on models with different client fractions to validate our hypothesis that small amounts of noise can produce large performance disparities by distorting sparse, yet critical cues.}
\ztdelete{We hypothesize that even minimal noise creates larger performance gaps with smaller client fractions by distorting sparse, critical information cues.}}

    \revision{
    \paragraph{Method.} 
    \ztedit{We compare models with different client fractions ($c=\{0.1,0.7\}$) with fixed $\epsilon$ values, and use SHAP and LIWC to analyze how the relationship between client fraction and privacy guarantees impacts learning markers of depression.}}

    \revision{
    \paragraph{Result.} \ztedit{We find that smaller client fractions are particularly susceptible to noise, severely distorting markers of depression. 
    At $c=0.1$ and $\epsilon=100$, the few depression markers are distorted in the training data, and the model relies primarily on irrelevant terms (e.g., ``Egypt'', ``witchcraft'', ``penguins'', ``ethanol'', ``fresco'') and a small set of genuine depression markers (e.g., ``withdrawal'', ``restraint'', ``escalated'') for classification. The identified LIWC categories in this low client fraction include percept/see (e.g., scans), relative/motion (e.g., removal), relative/space (e.g., eastwood, environments) -- all of which are potentially irrelevant markers of depression. While at $c=0.7$ and $\epsilon=100$, the top features include more markers of mental health conditions (e.g., health or bio terms: tumor, prescribed, intermittent; and social/work: advising, and religion: pilgirm). 
    Thus, we find that in considering the relationship between noise and classification for mental health conditions, we must take into account noise introduced, client fractions, and the density of relevant terms to our task.}}

    \ztdelete{Smaller client fractions with minimal noise severely distort depression markers. At 10\% client fraction ($\epsilon$=$100$), noise distorts the few depression cues in training data. With 70\% client fraction and similar noise, the model performs better due to more available information. At 10\% client fraction, top features were primarily unrelated terms (Egypt, witchcraft, Penguins, ethanol, fresco, surfer) with few genuine health cues (withdrawing, restraint, escalated, impacted). LIWC found no health categories and only isolated affect (magnificent) or negative emotion (witchcraft) terms. This demonstrates that limited training samples distort depression's sparse, subtle linguistic markers.}
    

\section{Implications} \label{implications}
We find that there is no statistical difference between models trained in centralized settings, which ensure no data privacy, and models trained using FL, which protect raw user data. 
\nnedit{However, our findings also show that differentially private FL results in a large utility drop when trying to mathematically guarantee privacy.}
Our findings indicate that FL can be a viable candidate for  \ztdelete{privacy-preserving ML} for the identification of high-risk behaviors, specifically the identification of people with depression and people who communicate suicidal ideation. 
\nnedit{\ztedit{In contrast, a}\ztdelete{A}dding noise to mathematically ensure privacy may not be a viable option, \ztedit{as adding noise has a large impact on model utility, and the negative impact of errors in mental health inference.} \ztdelete{given the large performance drop and negative impact of errors in mental health inference.}} Our findings have several implications.

First, our setup approximates a realistic, real-world setting by maintaining a non-identical data distribution, i.e., the token distributions (see \Cref{sec:token_dist}). Although IID data distribution across clients is common for FL research~\cite[e.g.,][]{Ji_Detecting_2019,galaFederatedApproachHate2023,gandhi_federated_2022}, creating IID settings would result in discarding relevant data, collecting multiple users on each client, or other bootstrapping approaches to ensure balanced datasets. 

Next, we consider two future implications of more private ML techniques for this tasks. 
\paragraph{Data Privacy and Dataset Sharing for Institutions:}
Privacy concerns prevent researchers (especially those without access to institutional review boards) from developing, accessing, and sharing mental health datasets. Our results indicate space for privacy-preserving approaches to data sharing, which can help address the privacy risks of sharing datasets. We consider the implications of this.

Researchers might build infrastructures that afford federated training approaches for datasets that are usually not shared due to ethics board restrictions. This approach leverages the client-server model of FL -- it allows institutions to retain control over local data and how it is shared while enabling other researchers to develop new predictive models for identifying high-risk behaviors and disorders. 
We are excited by the opportunity for future work.

\paragraph{Promoting More Consentful and User-Driven Data Sharing:}
Finally, we consider how researchers and practitioners may engineer more consentful and user-driven data sharing. Indeed, a major concern in ML and mental health research is the lack of explicit consent given by the data subjects for model training. \citet{pendse2024advancing} and \citet{ajmani2024data} discuss these concerns and suggest direct consent from users can be solicited to address this problem. However, centralized consent models are infeasible for a single institution to manage or gain for large datasets, and centralized ML infrastructure needs large datasets to demonstrate its prediction efficacy ~\cite{chancellor2019taxonomy}. 

As we show in this work, FL could involve opt-in data sharing and training from consenting users, while not decreasing performance on common tasks. 
Individuals could enroll in ML ecosystems, where they share model updates but keep their data on their own devices. Moreover, users could opt into model training for specific goals (triage, crisis intervention) rather than others they do not support (advertisements). A downstream impact of this consentful data sharing is that it could improve users' trust in mental health and AI technologies, which could support more adoption and potential applications. 
This must be done in ways that center users' perspectives and avoid harming them \cite{pendse2024advancing}.

\nnedit{Combining with the aforementioned degradation in performance to guarantee privacy,} we encourage the research community to develop infrastructures that allow users to opt in to research to identify high-risk behaviors and to create infrastructures that can safely share data with user consent.

\section{Conclusion}
\label{sec:conculsion}

This work provides a comprehensive analysis investigating the applicability of federated learning for the identification of high-risk behavior and disorders, spanning depressive disorder and suicidal ideation across two major platforms X and Reddit using benchmark datasets. 
We compare centralized approaches with federated learning settings \nnedit{and differentially private federated learning}, where users are simulated as client devices. 
We demonstrate that the federated approach performs comparably to the centralized method, and that performance differences for both tasks are statistically insignificant. \nnedit{Whereas there is a major utility-privacy trade-off when applying differentially private FL, which results in a large utility degradation when trying to mathematically guarantee the privacy. Training in smaller client fractions results in a larger utility drop-off.}
These findings open new avenues for mental health detection researchers to leverage federated learning, lowering data-sharing barriers that limit access to restricted social media mental health datasets and engaging in more consentful practices while adhering to the principle of preserving sensitive user data about their health and well-being. 


\section*{Ethical Considerations}
\label{sec:ethical_considerations}

Mental health behavior identification from social media has several ethical challenges.\ Inferring the mental status of users from their social media is risky; different actors can target individuals from these predictions. This may result in different harms, including exacerbating their mental health conditions. At the same time, such inference can also be beneficial. Early identification of these behaviors can pave the way to intervention. However, we must be aware of risks and engage in ethical practices suggested in prior work \cite{bentonEthicalResearchProtocols2017, chancellor2019taxonomy} as we deal with sensitive user social media data. 

The datasets used in our analysis are publicly available or accessed through IRB approval or data usage agreements to protect the users' privacy. The MTL and C-SSRS datasets were publicly available. The IRB institution at the University of Minnesota (Study ID: STUDY00022028) determined that our research does not involve human subjects, as we do not interact directly with the users. Hence, we accessed the other datasets through data use agreements, and shared this determination form through our requests for data use. Additionally, we ensured the data was only accessed by people included in the IRB determination form (the co-authors on this paper). Both the depression and STB datasets are anonymized and preprocessed to protect the identification of the users in future models. We applied further preprocessing techniques, such as cleaning any user mentions and URLs, to prevent these details from being included in training.

\section*{Limitations}
\label{sec:limitations}


For our analysis, we merged labels in the STB task into binary classifications of whether a user has STBs vs. not (see Datasets).\ \final{This collapsing of labels follows established prior work of both dataset creators:~\cite{shing-etal-2018-expert}, UMD-RD, collapse fine-grained risk levels to binary (at-risk vs control), and~\cite{gaur2019knowledge}, C-SSRS, who collapsed their five categories into 3 (where supportive and indicator classes are merged into one class: no-risk), for experimental purposes. However, it is important to note that binary labels may obscure clinically important distinctions, particularly whether DP noise disproportionately harms the detection of severe cases. Our goal in this work is to establish the viability of federated approaches before tackling fine-grained classifications.}

Furthermore, we studied depression on X data and STBs on Reddit data, which are the most commonly studied behaviors and platforms in the domain \cite{chancellor2020methods}. Our results do not yet extend to other conditions, such as anxiety, eating disorders, PTSD, or other social media platforms, such as Facebook, Weibo, or TikTok. Future work should examine other disorders and platforms.

\nnedit{Although federated learning \ztdelete{affords increased}\ztedit{presents improved} data privacy \ztedit{over centralized training paradigms} in terms of not sharing raw data, it is not a ``silver bullet'' for privacy concerns~\cite{jere2020taxonomy}, nor does it exempt researchers from making datasets to consider privacy as a core value in mental health and ML. Data could still be extracted from trained models~\cite{lyu2020threats}, and model performances can be degraded through model poisoning attacks~\cite{9308910}. 
Our findings show that when ensuring privacy by applying DP on FL, a large degradation in performance. It is therefore necessary for the community to investigate further mechanisms to ensure the data is protected from attacks ~\cite{9308910} such as homomorphic encryption~\cite{wen2023survey} and further experiments with differential privacy to realize the secure implementation of the proposed setting in the social media mental health detection domain.} \final{Other techniques such as parameter-efficient federated fine-tuning (LoRA/PEFT), personalized federated approaches, and analysis of the communication costs present future directions in this domain, building on the baselines presented by this work.} \revision{Furthermore, \ztedit{the relationship between federated learning, differential privacy, and fairness and bias in detecting mental health conditions remains under-explored and would require developing datasets with demographic details, and thus presents a rich avenue for future work}.
\ztdelete{future research could explore the fairness and bias of these privacy-preserving mechanisms. While the current datasets lack demographic details, investigating these factors remains a promising avenue for future work.}}

\bibliography{updated_references}
\bibliographystyle{acl_natbib}

\appendix

\section{Appendix}
\label{sec:appendix}

\subsection{Distribution of Tokens}
\label{sec:token_dist}

\begin{figure}[ht]
  \centering
  \includegraphics[width=0.5\textwidth]{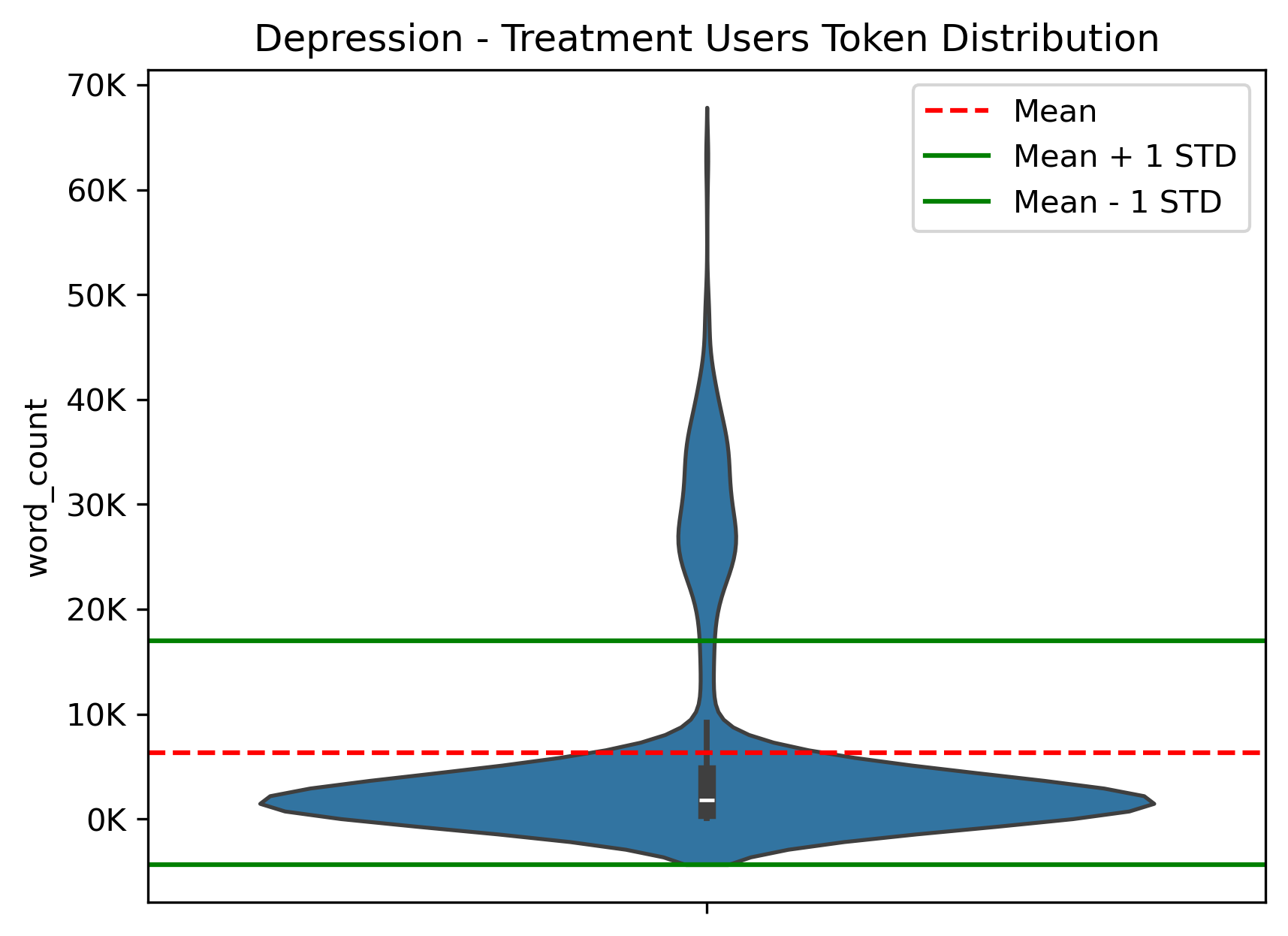}
 \caption{The violin plot illustrates the word count distributions of the treatment users in the depression detection dataset. Mean = $6324$, Standard Deviation = $10686$.}
 \label{fig_dep_treatment_tokens}
\end{figure}

\begin{figure}[ht]
  \centering
  \includegraphics[width=0.5\textwidth]{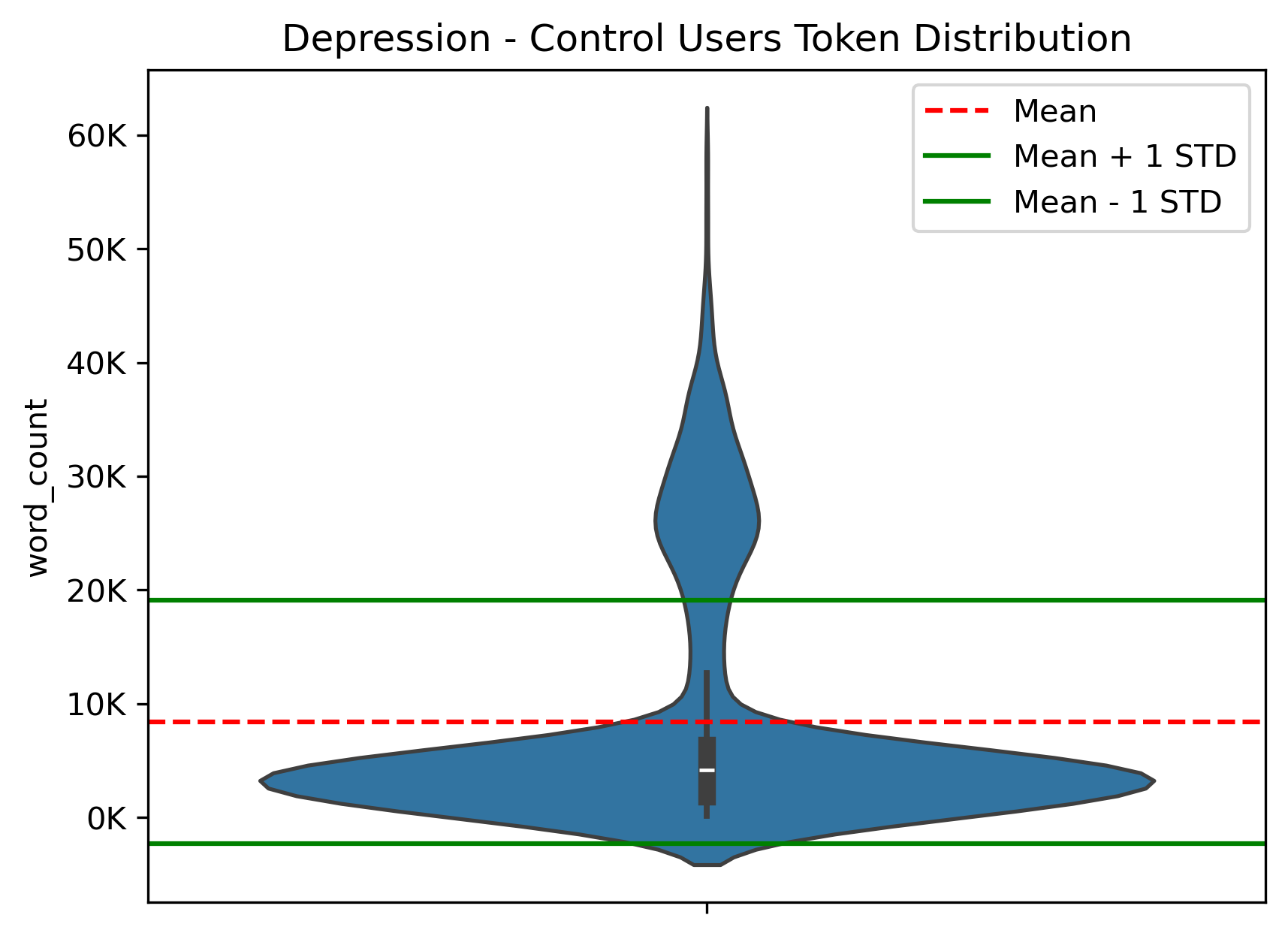}
 \caption{The violin plot illustrates the word count distributions of the control users in the depression detection dataset. Mean = $8402$, Standard Deviation = $10683$.}
 \label{fig_dep_control_tokens}
\end{figure}

\begin{figure}[ht]
  \centering
  \includegraphics[width=0.5\textwidth]{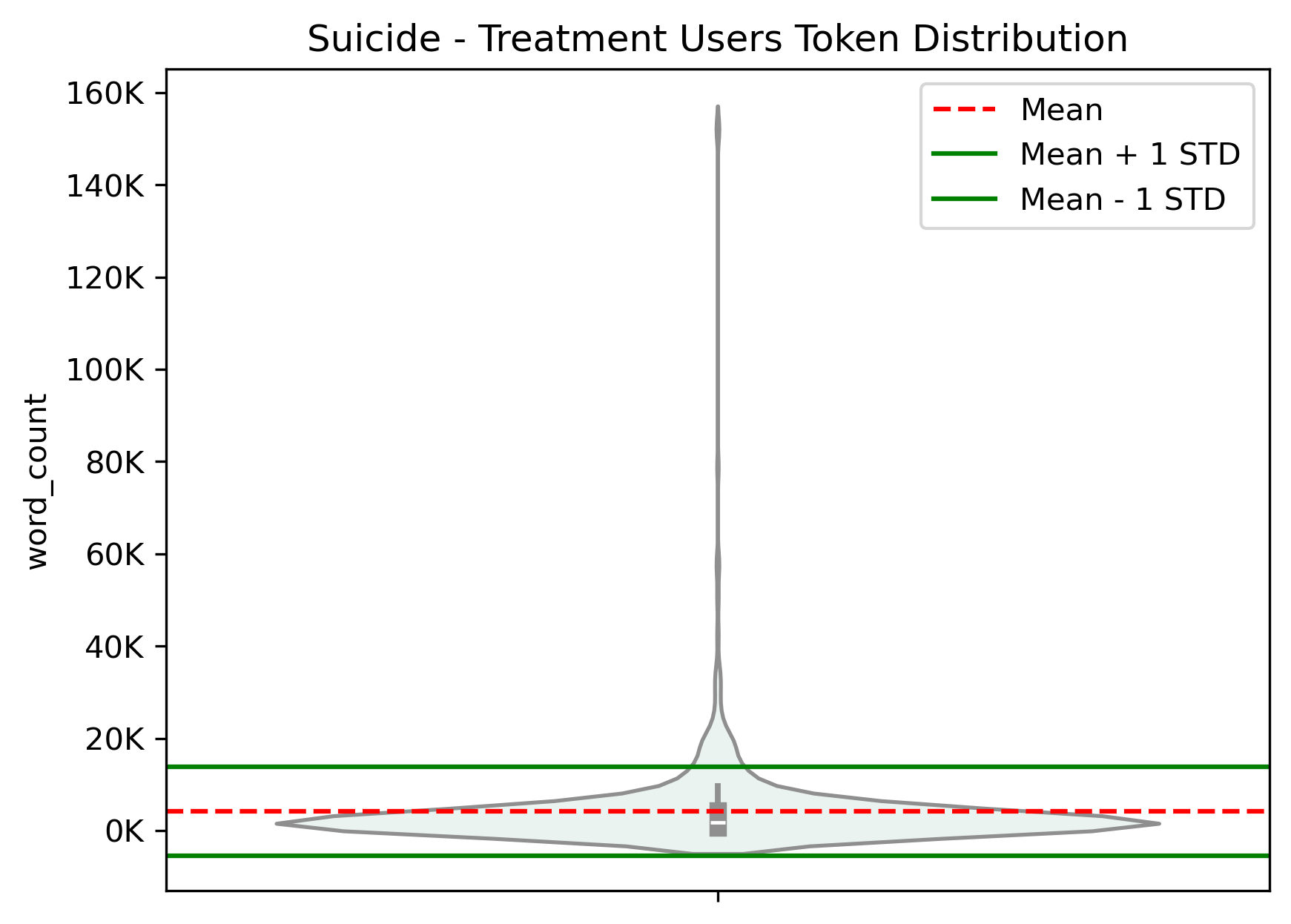}
 \caption{The violin plot illustrates the word count distributions of the treatment users in the suicidal thoughts behaviors identification dataset. Mean = $4282$, Standard Deviation = $10835$.}
 \label{fig_suicide_treatment_tokens}
\end{figure}

\begin{figure}[ht]
  \centering
  \includegraphics[width=0.5\textwidth]{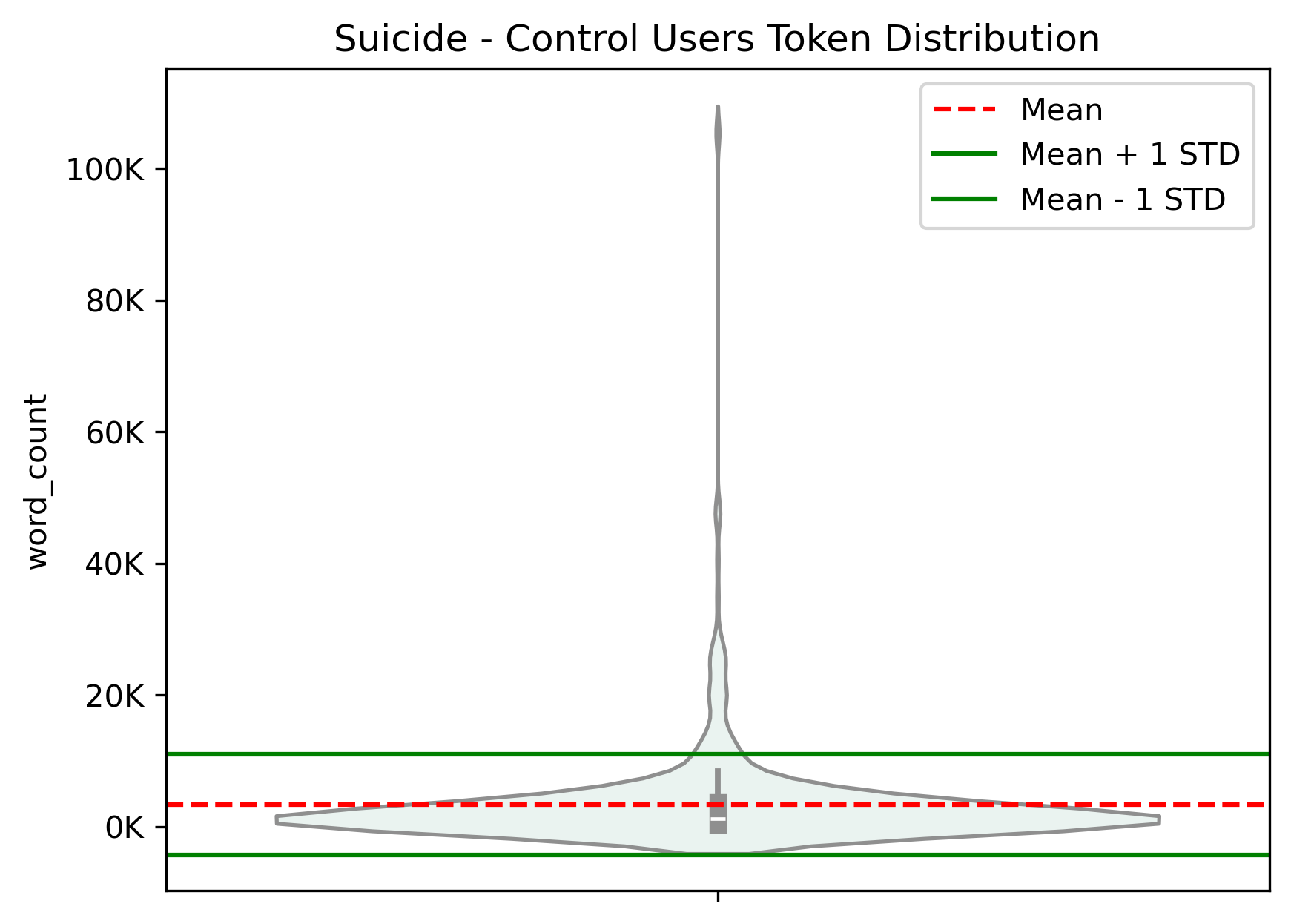}
 \caption{The violin plot illustrates the word count distributions of the control users in the suicidal thoughts behaviors identification dataset. Mean = $3378$, Standard Deviation = $7994$.}
 \label{fig_suicide_control_tokens}
\end{figure}

Here, we look at the token distribution within the aggregated depression and suicide datasets. There's a high standard deviation across the treatment and control users in both behaviors (depression and suicide).  In the depression dataset, the treatment users (STD=$10686$) have a slightly higher deviation from the control users (STD=$10683$).  
For the suicide dataset, the treatment users (STD=$10835$) have a higher standard deviation compared to the control ($7994$) counterparts.
This significant variation in distribution shows the non-Independent and Identical Distribution (non-IID) of data among the clients. This shows our simulation of unique individual users as clients depicting the real-world use case of high risk behavior identification.
(See \Cref{fig_dep_treatment_tokens,fig_dep_control_tokens} for treatment and control depression users, respectively. \Cref{fig_suicide_treatment_tokens,fig_suicide_control_tokens} for treatment and control suicide users, respectively)

\subsection{Model Parameters and Experimental Setup}
\label{app:sec:replication}

The learning rate is set to $4e-5$. Adam is used as an optimizer. 
We trained for `epochs = $50$' and applied an early stopping of $5$. 
A batch size of $32$ is used for all the models except for MentalLongformer, where we used 16 due to memory limits. We used the default Huggingface parameter values for the remaining hyperparameters for both centralized and federated settings. 

\nnedit{For our \textbf{Differentially Private Federated Learning (DP-FL)} setup, we used finetuned BERT on the depression identification task. We set the clipping norm ($clip\_norm = 1$), the privacy budget is experimented ($\epsilon$ = 1, 5, 10, 50, 100), we used the fedprox aggregation algorithm, client fraction (client\_fraction = {0.1, 0.7}, 10\% and 70\% of the data, respectively). Learning rate ($client\_lr = 4e-5$, $server\_lr = 1e-3$) provided the best performance after searching over higher learning rate space ($client\_lr = 1e-2$, $server\_lr = 1e-1$). Training for `$epochs = 50$' and `$rounds = 100$' yielded the best result (search space `$epochs = 50, 100$' and `$rounds = 50, 100, 200$'), and applied an early stopping of $5$, with batch\_size of 32. We used delta ($\delta$ = $1e-5$). The sigma is calculated with an adaptive Gaussian mechanism.  } 

\section{Depression and Suicidal Thoughts and Behaviors Detection Results}
\label{app:sec:dep_suicidal_results}

\begin{table*}[h!]
\centering
    \small 
    \resizebox{1.0\textwidth}{!} {
\begin{tabular}{@{}l*{14}{c}@{}}
        \toprule
        \multirow{2}{*}{\textbf{Client Fraction}} & \multicolumn{2}{c}{\textbf{Logistic Regression}} & \multicolumn{2}{c}{\textbf{MentalBERT}} & \multicolumn{2}{c}{\textbf{MentalLongformer}} & \multicolumn{2}{c}{\textbf{BERT}} & \multicolumn{2}{c}{\textbf{RoBERTa}} & \multicolumn{2}{c}{\textbf{DistilBERT}} & \multicolumn{2}{c}{\textbf{DistilRoBERTa}} \\
        \cmidrule(lr){2-3} \cmidrule(lr){4-5} \cmidrule(lr){6-7} \cmidrule(lr){8-9} \cmidrule(lr){10-11} \cmidrule(lr){12-13} \cmidrule(l){14-15}
        & Recall & F1 & Recall & F1 & Recall & F1 & Recall & F1 & Recall & F1 & Recall & F1 & Recall & F1 \\
        \midrule
        \textbf{c = 10\%} & 59.39 & 59.39 & 76.89 & 76.85 & \textbf{82.99} & \textbf{83.16} & 73.96 & 73.93 & 75.06 & 75.00 & 74.25 & 74.25 & 72.89 & 72.98 \\
        \addlinespace
        \textbf{c = 30\%} & 55.47 & 55.46 & 77.02 & 77.01 & 82.29 & 82.40 & 74.06 & 73.48 & 75.30 & 75.21 & 74.15 & 74.20 & 73.31 & 73.37 \\
        \addlinespace
        \textbf{c = 50\%} & 56.05 & 56.05 & 76.62 & 76.59 & 81.98 & 82.07 & 76.64 & 76.40 & \textbf{76.85} & \textbf{76.83} & 75.12 & 75.13 & 73.81 & 73.85 \\
        \addlinespace
        \textbf{c = 70\%} & 55.95 & 55.94 & 75.57 & 75.34 & 82.48 & 82.63 & 77.03 & 76.83 & 75.92 & 75.88 & 75.29 & 75.37 & 74.18 & 74.21 \\
        \bottomrule
    \end{tabular}
}
\caption{Federated approach on the \textbf{\texttt{depression dataset}} using \textbf{\texttt{fedAvg}} algorithm in different client fractions.}
    \label{tab:fedavg_depression_results}
\end{table*}

\begin{table*}[h!]
\centering
    \small 
    \resizebox{1.0\textwidth}{!} {
\begin{tabular}{@{}l*{14}{c}@{}}
        \toprule
        \multirow{2}{*}{\textbf{Client Fraction}} & \multicolumn{2}{c}{\textbf{Logistic Regression}} & \multicolumn{2}{c}{\textbf{MentalBERT}} & \multicolumn{2}{c}{\textbf{MentalLongformer}} & \multicolumn{2}{c}{\textbf{BERT}} & \multicolumn{2}{c}{\textbf{RoBERTa}} & \multicolumn{2}{c}{\textbf{DistilBERT}} & \multicolumn{2}{c}{\textbf{DistilRoBERTa}} \\
        \cmidrule(lr){2-3} \cmidrule(lr){4-5} \cmidrule(lr){6-7} \cmidrule(lr){8-9} \cmidrule(lr){10-11} \cmidrule(lr){12-13} \cmidrule(l){14-15}
        & Recall & F1 & Recall & F1 & Recall & F1 & Recall & F1 & Recall & F1 & Recall & F1 & Recall & F1 \\
        \midrule
        \textbf{c = 10\%} & 55.91 & 55.90 & \textbf{77.25} & \textbf{77.08} & 78.19 & 78.35 & \textbf{77.62} & \textbf{77.57} & 76.69 & 76.71 & 74.78 & 74.82 & 73.34 & 73.06 \\
        \addlinespace
        \textbf{c = 30\%} & 55.05 & 54.90 & 75.29 & 75.00 & 80.02 & 80.11 & 76.75 & 76.41 & 76.14 & 76.08 & 74.91 & 74.78 & 73.81 & 73.76 \\
        \addlinespace
        \textbf{c = 50\%} & 55.77 & 55.78 & 75.81 & 75.60 & 81.72 & 81.80 & 76.16 & 75.79 & 75.41 & 75.41 & 75.33 & 75.37 & 73.67 & 73.75 \\
        \addlinespace
        \textbf{c = 70\%} & 55.76 & 55.75 & 75.71 & 75.51 & 79.93 & 80.06 & 76.37 & 76.13 & 75.73 & 75.65 & 74.57 & 74.55 & 73.24 & 73.31 \\
        \bottomrule
    \end{tabular}
}
\caption{Federated approach on the \textbf{\texttt{depression dataset}} using \textbf{\texttt{fedProx}} algorithm in different client fractions.}
    \label{tab:fedprox_depression_results}
\end{table*}

\begin{table*}[h!]
\centering
    \small 
    \resizebox{1.0\textwidth}{!} {
\begin{tabular}{@{}l*{14}{c}@{}}
        \toprule
        \multirow{2}{*}{\textbf{Client Fraction}} & \multicolumn{2}{c}{\textbf{Logistic Regression}} & \multicolumn{2}{c}{\textbf{MentalBERT}} & \multicolumn{2}{c}{\textbf{MentalLongformer}} & \multicolumn{2}{c}{\textbf{BERT}} & \multicolumn{2}{c}{\textbf{RoBERTa}} & \multicolumn{2}{c}{\textbf{DistilBERT}} & \multicolumn{2}{c}{\textbf{DistilRoBERTa}} \\
        \cmidrule(lr){2-3} \cmidrule(lr){4-5} \cmidrule(lr){6-7} \cmidrule(lr){8-9} \cmidrule(lr){10-11} \cmidrule(lr){12-13} \cmidrule(l){14-15}
        & Recall & F1 & Recall & F1 & Recall & F1 & Recall & F1 & Recall & F1 & Recall & F1 & Recall & F1 \\
        \midrule
        \textbf{c = 10\%} & 66.90 & 66.83 & 70.86 & 69.96 & 74.10 & 74.04 & 71.38 & 71.28 & 71.96 & 71.80 & \textbf{77.33} & \textbf{77.42} & 68.88 & 67.74 \\
        \addlinespace
        \textbf{c = 30\%} & 68.82 & 68.84 & 74.61 & 74.44 & 78.25 & 77.62 & 76.02 & 75.98 & 66.89 & 63.38 & 76.78 & 76.70 & 76.78 & 76.94 \\
        \addlinespace
        \textbf{c = 50\%} & \textbf{69.03} & \textbf{69.06} & 69.84 & 68.87 & 64.87 & 59.84 & 75.16 & 75.15 & 68.40 & 66.88 & 74.58 & 74.12 & 78.00 & 78.11 \\
        \addlinespace
        \textbf{c = 70\%} & 68.50 & 68.56 & 70.61 & 70.41 & 82.64 & 82.75 & 64.63 & 59.85 & 75.21 & 75.29 & 76.36 & 76.41 & \textbf{79.63} & \textbf{79.73} \\
        \bottomrule
    \end{tabular}
}
\caption{Federated approach on the \textbf{\texttt{depression dataset}} using \textbf{\texttt{fedOpt}} algorithm in different client fractions.}
    \label{tab:fedopt_depression_results}
\end{table*}



\begin{table*}[h!]
\centering
    \small 
    \resizebox{1.0\textwidth}{!} {
\begin{tabular}{@{}l*{14}{c}@{}}
        \toprule
        \multirow{2}{*}{\textbf{Client Fraction}} & \multicolumn{2}{c}{\textbf{Logistic Regression}} & \multicolumn{2}{c}{\textbf{MentalBERT}} & \multicolumn{2}{c}{\textbf{MentalLongformer}} & \multicolumn{2}{c}{\textbf{BERT}} & \multicolumn{2}{c}{\textbf{RoBERTa}} & \multicolumn{2}{c}{\textbf{DistilBERT}} & \multicolumn{2}{c}{\textbf{DistilRoBERTa}} \\
        \cmidrule(lr){2-3} \cmidrule(lr){4-5} \cmidrule(lr){6-7} \cmidrule(lr){8-9} \cmidrule(lr){10-11} \cmidrule(lr){12-13} \cmidrule(l){14-15}
        & Recall & F1 & Recall & F1 & Recall & F1 & Recall & F1 & Recall & F1 & Recall & F1 & Recall & F1 \\
        \midrule
        \textbf{c = 10\%} & 59.15 & 58.95 & 71.84 & 70.96 & 77.14 & 77.05 & 75.68 & 75.59 & 80.51 & 80.65 & 74.32 & 74.31 & 73.89 & 73.79 \\
        \addlinespace
        \textbf{c = 30\%} & 58.56 & 58.33 & 82.87 & 83.05 & 79.41 & 79.54 & 78.52 & 78.56 & \textbf{81.51} & \textbf{81.62} & 73.27 & 73.15 & 68.90 & 67.38 \\
        \addlinespace
        \textbf{c = 50\%} & 57.91 & 57.71 & \textbf{83.95} & \textbf{84.09} & \textbf{81.56} & \textbf{81.73} & 80.15 & 80.27 & 79.89 & 79.88 & 74.77 & 74.66 & 79.14 & 79.20 \\
        \addlinespace
        \textbf{c = 70\%} & 58.89 & 58.64 & 81.96 & 82.12 & 81.46 & 81.63 & \textbf{82.65} & \textbf{82.80} & 77.14 & 77.05 & 75.49 & 75.48 & 72.17 & 71.34 \\
        \bottomrule
    \end{tabular}
}
\caption{Federated approach on the \textbf{\texttt{suicide dataset}} using \textbf{\texttt{fedAvg}} algorithm in different client fractions.}
    \label{tab:fedavg_suicide_results}
\end{table*}

\begin{table*}[h!]
\centering
    \small 
    \resizebox{1.0\textwidth}{!} {
\begin{tabular}{@{}l*{14}{c}@{}}
        \toprule
        \multirow{2}{*}{\textbf{Client Fraction}} & \multicolumn{2}{c}{\textbf{Logistic Regression}} & \multicolumn{2}{c}{\textbf{MentalBERT}} & \multicolumn{2}{c}{\textbf{MentalLongformer}} & \multicolumn{2}{c}{\textbf{BERT}} & \multicolumn{2}{c}{\textbf{RoBERTa}} & \multicolumn{2}{c}{\textbf{DistilBERT}} & \multicolumn{2}{c}{\textbf{DistilRoBERTa}} \\
        \cmidrule(lr){2-3} \cmidrule(lr){4-5} \cmidrule(lr){6-7} \cmidrule(lr){8-9} \cmidrule(lr){10-11} \cmidrule(lr){12-13} \cmidrule(l){14-15}
        & Recall & F1 & Recall & F1 & Recall & F1 & Recall & F1 & Recall & F1 & Recall & F1 & Recall & F1 \\
        \midrule
        \textbf{c = 10\%} & 59.15 & 58.95 & 75.97 & 75.88 & 69.20 & 67.66 & 78.38 & 78.48 & 78.88 & 78.92 & 74.29 & 74.26 & \textbf{81.63} & \textbf{81.79} \\
        \addlinespace
        \textbf{c = 30\%} & 58.57 & 58.33 & 82.35 & 82.50 & 70.93 & 70.01 & 78.00 & 78.07 & 81.44 & 81.58 & 73.20 & 73.02 & 70.54 & 69.39 \\
        \addlinespace
        \textbf{c = 50\%} & 57.91 & 57.72 & 81.53 & 81.70 & 74.66 & 74.46 & 76.88 & 76.82 & 80.67 & 80.78 & 73.09 & 72.80 & 77.89 & 77.92 \\
        \addlinespace
        \textbf{c = 70\%} & 58.89 & 58.64 & 82.87 & 83.05 & 79.10 & 79.15 & 79.82 & 79.93 & 79.91 & 80.03 & 73.20 & 73.16 & 77.15 & 77.05 \\
        \bottomrule
    \end{tabular}
}
\caption{Federated approach on the \textbf{\texttt{suicide dataset}} using \textbf{\texttt{fedProx}} algorithm in different client fractions.}
    \label{tab:fedprox_suicide_results}
\end{table*}

\begin{table*}[h!]
\centering
    \small 
    \resizebox{1.0\textwidth}{!} {
\begin{tabular}{@{}l*{14}{c}@{}}
        \toprule
        \multirow{2}{*}{\textbf{Client Fraction}} & \multicolumn{2}{c}{\textbf{Logistic Regression}} & \multicolumn{2}{c}{\textbf{MentalBERT}} & \multicolumn{2}{c}{\textbf{MentalLongformer}} & \multicolumn{2}{c}{\textbf{BERT}} & \multicolumn{2}{c}{\textbf{RoBERTa}} & \multicolumn{2}{c}{\textbf{DistilBERT}} & \multicolumn{2}{c}{\textbf{DistilRoBERTa}} \\
        \cmidrule(lr){2-3} \cmidrule(lr){4-5} \cmidrule(lr){6-7} \cmidrule(lr){8-9} \cmidrule(lr){10-11} \cmidrule(lr){12-13} \cmidrule(l){14-15}
        & Recall & F1 & Recall & F1 & Recall & F1 & Recall & F1 & Recall & F1 & Recall & F1 & Recall & F1 \\
        \midrule
        \textbf{c = 10\%} & \textbf{65.24} & \textbf{65.12} & 72.46 & 72.41 & 54.58 & 43.93 & 70.14 & 69.88 & 67.73 & 66.30 & \textbf{77.81} & \textbf{77.75} & 73.05 & 72.46 \\
        \addlinespace
        \textbf{c = 30\%} & 64.86 & 64.48 & 76.44 & 76.45 & 68.51 & 68.54 & 65.94 & 64.55 & 70.74 & 70.73 & 69.26 & 68.78 & 72.36 & 72.23 \\
        \addlinespace
        \textbf{c = 50\%} & 62.68 & 62.32 & 76.20 & 74.96 & 62.52 & 60.80 & 71.24 & 70.99 & 70.85 & 70.87 & 69.26 & 68.78 & 69.85 & 69.60 \\
        \addlinespace
        \textbf{c = 70\%} & 62.50 & 62.34 & 77.77 & 77.74 & 57.36 & 50.45 & 63.16 & 59.74 & 74.10 & 74.04 & 74.94 & 74.93 & 69.85 & 69.60 \\
        \bottomrule
    \end{tabular}
}
\caption{Federated approach on the \textbf{\texttt{suicide dataset}} using \textbf{\texttt{fedOpt}} algorithm in different client fractions.}
    \label{tab:fedopt_suicide_results}
\end{table*}

\end{document}